\documentclass[11pt,a4paper]{article}
\usepackage[hyperref]{acl2021}
\usepackage{times}
\usepackage{latexsym}

\usepackage{microtype}
\usepackage{makecell}
\usepackage{multirow}
\usepackage{amsfonts} 
\usepackage{amsmath} 
\usepackage{adjustbox}
\usepackage{xpatch}  
\usepackage{arydshln} 
\def\aclpaperid{1771} 
\newcommand\numberthis{\addtocounter{equation}{1}\tag{\theequation}}

\xpretocmd{\footnote}{\unskip}{}{}
\usepackage{xurl}

\aclfinalcopy

\title{Confidence-Aware Scheduled Sampling for \\ Neural Machine Translation}
\author{
  Yijin Liu\textsuperscript{1}\thanks{\ \ This work was done when Yijin Liu was interning at Pattern Recognition Center, WeChat AI, Tencent Inc, China} \ ,
  Fandong Meng\textsuperscript{2}, 
  Yufeng Chen\textsuperscript{1} 
  Jinan Xu\textsuperscript{1}\thanks{ \ \ Jinan Xu is the corresponding author of the paper.} ,
  and Jie Zhou\textsuperscript{2} \\
  \textsuperscript{1}Beijing Jiaotong University, China \\
  \textsuperscript{2}Pattern Recognition Center, WeChat AI, Tencent Inc, China \\
  \texttt{adaxry@gmail.com} \\
  \texttt{\{fandongmeng, withtomzhou\}@tencent.com} \\
  \texttt{\{jaxu,chenyf\}@bjtu.edu.cn} \\
}

\date{}

\begin{document}
\maketitle

\begin{abstract}
Scheduled sampling is an effective method to alleviate the exposure bias problem of neural machine translation. It simulates the inference scene by randomly replacing ground-truth target input tokens with predicted ones during training. Despite its success, its critical schedule strategies are merely based on training steps, ignoring the real-time model competence, which limits its potential performance and convergence speed. To address this issue, we propose confidence-aware scheduled sampling. Specifically, we quantify real-time model competence by the confidence of model predictions, based on which we design fine-grained schedule strategies. In this way, the model is exactly exposed to predicted tokens for high-confidence positions and still ground-truth tokens for low-confidence positions. Moreover, we observe vanilla scheduled sampling suffers from degenerating into the original teacher forcing mode since most predicted tokens are the same as ground-truth tokens. Therefore,  under the above confidence-aware strategy, we further expose more noisy tokens ({\em e.g.,} wordy and incorrect word order) instead of predicted ones for high-confidence token positions. We evaluate our approach on the Transformer and conduct experiments on large-scale WMT 2014 English-German, WMT 2014 English-French, and WMT 2019 Chinese-English. Results show that our approach significantly outperforms the Transformer and vanilla scheduled sampling on both translation quality and convergence speed.
\end{abstract}

\section{Introduction}
\label{sec:introduction}

Neural Machine Translation (NMT) has made promising progress in recent years \cite{seq2seq_2014,bahdanau_nmt_2015,transformer_2017}. Generally, NMT models are trained to maximize the likelihood of the next token given previous golden tokens as inputs, {\em i.e.,} teacher forcing \cite{teacher_forcing_2016}. 
However, at the inference stage, golden tokens are unavailable. The model is exposed to an unseen data distribution generated by itself.
This discrepancy between training and inference is named as the \textit{exposure bias} problem \cite{mixer_2015}. 

Many techniques have been proposed to alleviate the exposure bias problem. 
To our knowledge, they mainly fall into two categories. The one is sentence-level training, which treats the sentence-level metric ({\em e.g.,} BLEU) as a reward, and directly maximizes the expected rewards of generated sequences \cite{mixer_2015,mrt_2016,rl_related_2017}.
Although intuitive, they generally suffer from slow and unstable training due to the high variance of policy gradients and the credit assignment problem \cite{credit_assignment_1985,rl_related1_2018,rl_related2_2018}.
Another category is sampling-based approaches, aiming to simulate the data distribution of reference during training. Scheduled sampling \cite{ss_2015} is a representative method, which samples tokens between golden references and model predictions with a scheduled probability. 
\citet{zhang_bridging_2019} further refine the sampling space of scheduled sampling with predictions from beam search.
\citet{ss_transformer_2019} and \citet{parallel_ss_2019} extend scheduled sampling to the Transformer with a novel two-pass decoding architecture.

Although these sampling-based approaches have been shown effective, most of them schedule the sampling probability based on training steps.
We argue this schedule strategy has two following limitations: 
1) It is far from exactly reflecting the real-time model competence; 2) It is only based on training steps and equally treat all token positions, which is too coarse-grained to guide the sampling selection for each target token.
These two limitations yield an inadequate and inefficient schedule strategy, which hinders the potential performance and convergence speed of vanilla scheduled sampling-based approaches.

To address these issues, we propose confidence-aware scheduled sampling. 
Specifically, we take the model prediction confidence as the assessment of real-time model competence, based on which we design fine-grained schedule strategies.
Namely, we sample predicted tokens as target inputs for high-confidence positions and still ground-truth tokens for low-confidence positions.
In this way, the NMT model is exactly exposed to corresponding tokens according to its real-time competence rather than coarse-grained predefined patterns.
Additionally, we observe that most predicted tokens are the same as ground-truth tokens due to teacher forcing\footnote{We observe that about 70\% tokens are correctly predicted in WMT14 EN-DE training data.}, degenerating scheduled sampling to the original teacher forcing mode. 
Therefore, we further expose more noisy tokens \cite{wechat_wmt_2020} ({\em e.g.,} wordy and incorrect word order) instead of predicted ones for high-confidence token positions.
Experimentally, we evaluate our approach on the Transformer \cite{transformer_2017} and conduct experiments on large-scale WMT 2014 English-German (EN-DE), WMT 2014 English-French  (EN-FR), and WMT 2019 Chinese-English  (ZH-EN).


The main contributions of this paper can be summarized as follows\footnote{Codes are  available at \url{https://github.com/Adaxry/conf_aware_ss4nmt}.}:
\begin{itemize}
    \item To the best of our knowledge, we are the first to propose confidence-aware scheduled sampling for NMT, which exactly samples corresponding tokens according to the real-time model competence rather than coarse-grained predefined patterns.
    \item We further explore to sample more noisy tokens for high-confidence token positions, preventing scheduled sampling from degenerating into the original teacher forcing mode.
    \item Our approach significantly outperforms the Transformer by 1.01, 1.03, 0.98 BLEU and outperforms the stronger scheduled sampling by 0.51, 0.41, and 0.58 BLEU  on EN-DE, EN-FR, and ZH-EN, respectively.
    Our approach speeds up model convergence about 3.0$\times$ faster than the Transformer and about 1.8$\times$ faster than vanilla scheduled sampling.
    \item Extensive analyses indicate the effectiveness and superiority of our approach on longer sentences. Moreover, our approach can facilitate the training of the Transformer model with deeper decoder layers.
\end{itemize}

\begin{figure}[t!]
\begin{center}
     \scalebox{0.47}{
      \includegraphics[width=1\textwidth]{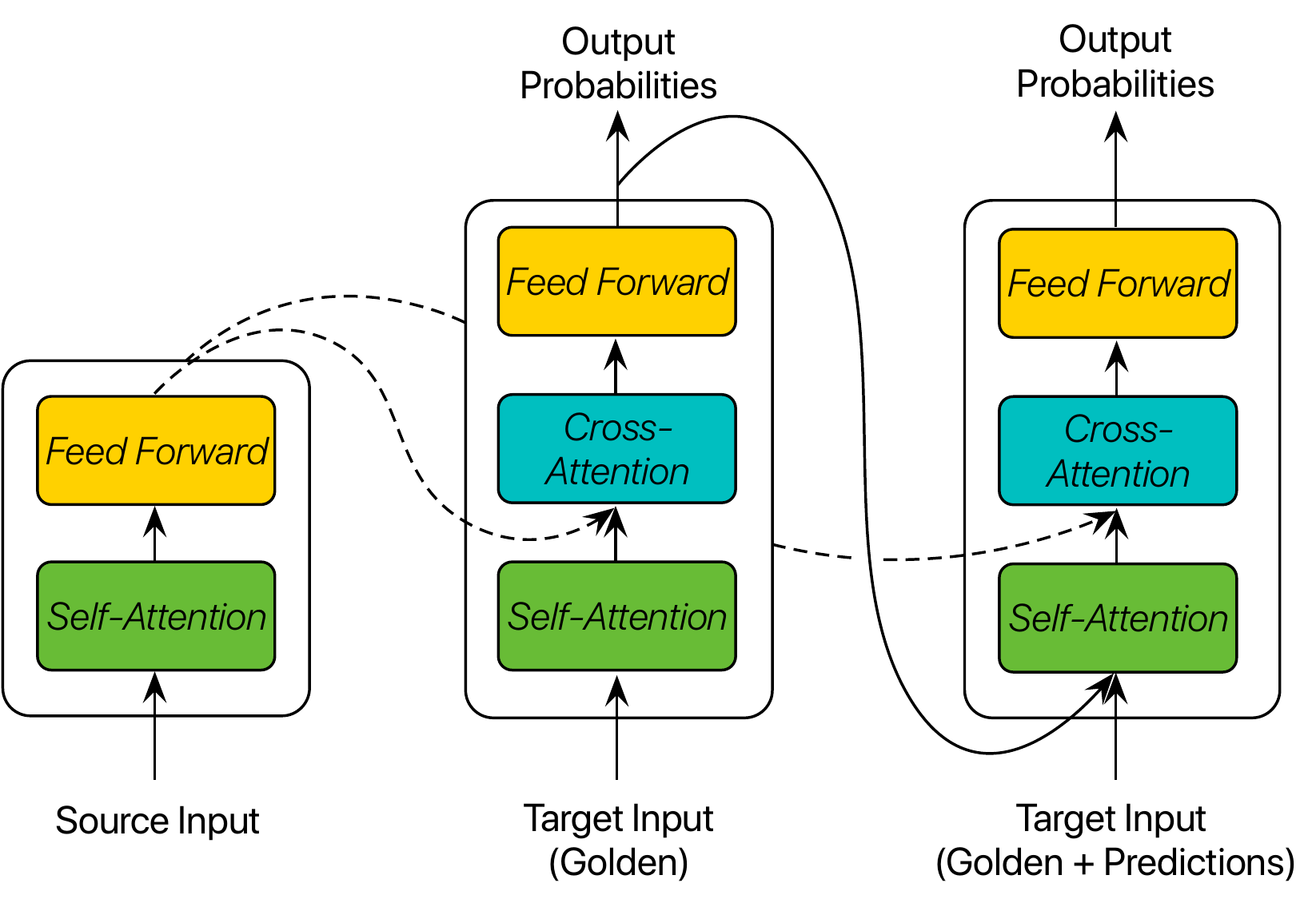}
      } 
      \caption{
      Scheduled sampling for the transformer with two-pass decoding \cite{ss_transformer_2019}.
      } 
      \label{fig:ss_for_transformer}  
 \end{center} 
\end{figure}

\section{Background}
\subsection{Neural Machine Translation}
Given a pair of source language $\mathbf{X} = \{x_1, x_2, \cdots, x_m \}$ with $m$ tokens and target language $\mathbf{Y} = \{y_1, y_2, \cdots, y_n \}$ with $n$ tokens, neural machine translation aims to model the following translation probability:
\begin{align*}
     P(\mathbf{Y}|\mathbf{X}) 
     & = \prod_{t=1}^{n}{P(y_{t} | \mathbf{y}_{<t}, \mathbf{X}, \theta)} \\
     & = \sum_{t=1}^{n}{\log P(y_{t} | \mathbf{y}_{<t}, \mathbf{X}, \theta)}
     \numberthis
     \label{equ:nmt_define}
\end{align*}    
where $t$ is the index of target tokens, $\mathbf{y}_{<t}$ is the partial translation before $y_t$, and $\theta$ is model parameter. In the training stage, $\mathbf{y}_{<t}$ are ground-truth tokens, and this procedure is also known as teacher forcing. The translation model is generally trained with maximum likelihood estimation (MLE).

\subsection{Scheduled Sampling for the Transformer}
 Scheduled sampling is initially designed for Recurrent Neural Networks \cite{ss_2015}, and further modifications are needed when applied to the Transformer \cite{ss_transformer_2019,parallel_ss_2019}.
As shown in Figure \ref{fig:ss_for_transformer}, we follow the two-pass decoding architecture. 
In the first pass, the model conducts the same as a standard NMT model. Its predictions are used to simulate the inference scene \footnote{Following \citeauthor{prediction_softmax_2017} (\citeyear{prediction_softmax_2017}), model predictions are the weighted sum of target embeddings over output probabilities. As model predictions cause a mismatch with golden tokens, they can simulate translation errors of the inference scene.}.
In the second pass, inputs of the decoder $\widetilde{\mathbf{y}}_{<t}$ are sampled from predictions of the first pass and ground-truth tokens with a certain probability. 
Finally, predictions of the second pass are used to calculate the cross-entropy loss, and Equation (\ref{equ:nmt_define}) is modified as follow:
\begin{align*}
     P(\mathbf{Y}|\mathbf{X}) = \sum_{t=1}^{n}{log P(y_{t} | \widetilde{\mathbf{y}}_{<t}, \mathbf{X}, \theta)}
     \numberthis
     \label{equ:ss_for_transformer}
\end{align*}    
Note that the two decoders are identical and share the same parameters. At inference, only the first decoder is used, that is just the standard Transformer. How to schedule the above probability of sampling tokens is the key point, which is exactly what we aim to improve in this paper. 

\subsection{Decay Strategies on Training Steps}
\label{sec:decay_strategies_on_training_steps}
\begin{figure}[t!]
\begin{center}
     \scalebox{0.5}{
      \includegraphics[width=1\textwidth]{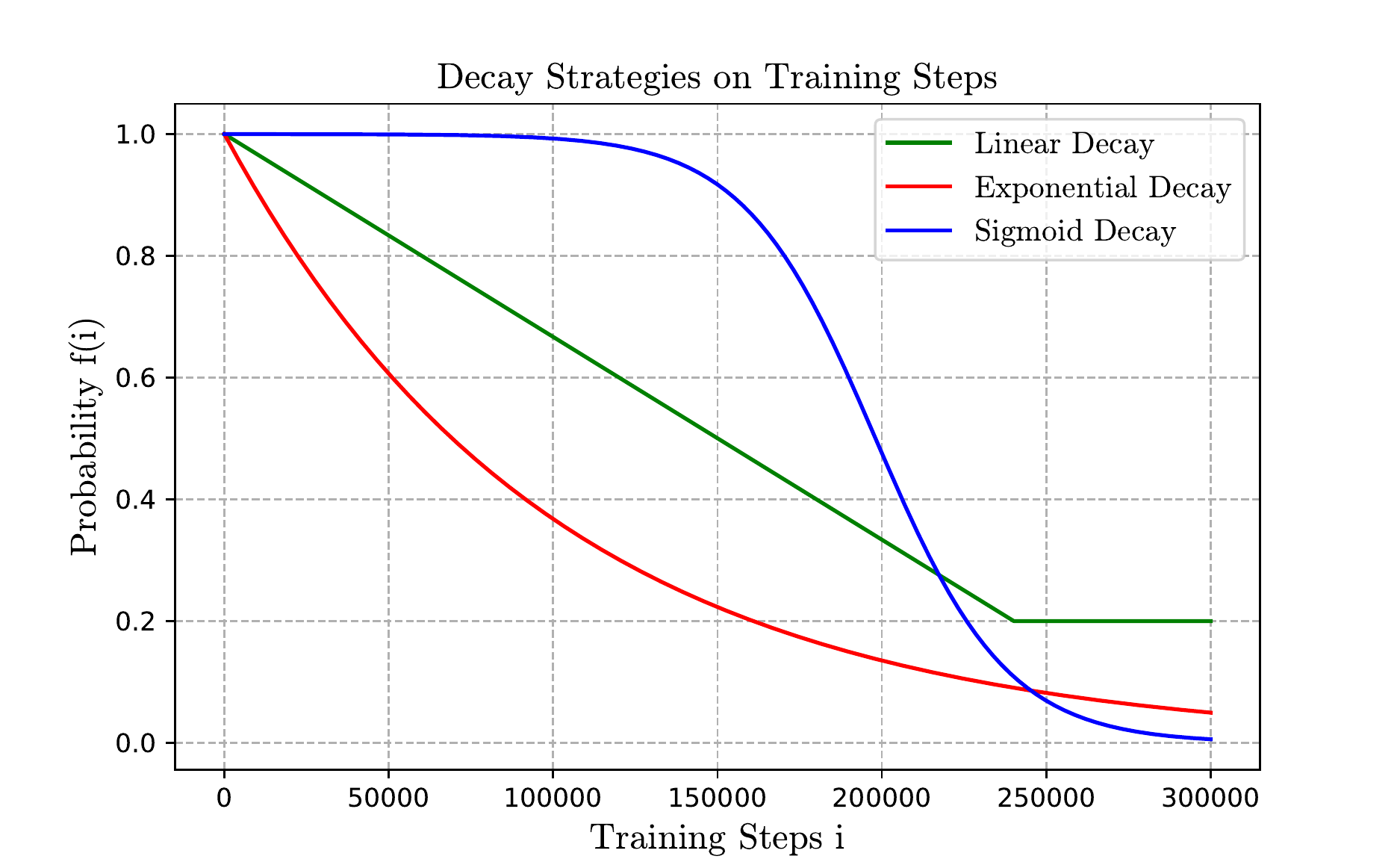}
      } 
      \caption{
      Examples of different decay strategies $f(i)$.
      } 
      \label{fig:training_step_decay_strategy}  
 \end{center} 
\end{figure}

Existing schedule strategies are based on training steps \cite{ss_2015,zhang_bridging_2019}. As the number of the training step $i$ increases, the model should be exposed to its own predictions more frequently.  At the $i$-th training step, the probability of sampling golden tokens $f(i)$ is calculated as follow:
\begin{itemize}
\item Linear Decay: $f(i) = \max(\epsilon , ki + b)$, where $\epsilon$ is the minimum value, and $k < 0$ and $b$ are respectively the slope and offset of the decay.
\item Exponential Decay: $f(i) = k^i$, where $k < 1$ is the radix to adjust the sharpness of the decay.
\item Inverse Sigmoid Decay: $f(i) = \frac{k}{k + e^{\frac{i}{k}}}$, where $e$ is the mathematical constant, and $k \geq 1$ is a hyperparameter to adjust the sharpness of the decay.
\end{itemize}
We draw visible examples for different decay strategies in Figure \ref{fig:training_step_decay_strategy}.

\section{Approaches}
In this section, we firstly describe how to estimate model confidence at each token position. Secondly, we elaborate the fine-grained schedule strategy based on model confidence. Finally, we explore to sample more noisy tokens instead of predicted tokens for high-confidence positions.

\subsection{Model Confidence Estimation}
\label{sec:model_confidence_estimation}
We explore two approaches to estimate model confidence at each token position.

\paragraph{Predicted Translation Probability (PTP).}
Current NMT models are well-calibrated with regularization techniques in the training setting \cite{analyze_uncertainty_nmt_2018,calibrated_2019,calibration_acl_2020}.
Namely the predicted translation probability can directly serve as the model confidence. At the $t$-th target token position, we calculate the model confidence $conf(t)$ as follow:
\begin{align}
conf(t) = P(y_{t} | \mathbf{y}_{<t}, \mathbf{X}, {\theta})
\numberthis
\label{equ:ptp}
\end{align}
Since we base our approach on the Transformer with two-pass decoding \cite{ss_transformer_2019,parallel_ss_2019}, above predicted translation probability can be directly obtained in the first-pass decoding (shown in Figure \ref{fig:ss_for_transformer}), causing no additional computation costs.

\paragraph{Monte Carlo Dropout Sampling.}


The model confidence can be quantified by Bayesian neural networks \cite{bayesian_1991,bayesian_2012}, which place distributions over the weights of neural networks. 
We adopt widely used Monte Carlo dropout sampling \cite{mc_dropout_2016,uncertainty_bt_2019} to approximate Bayesian inference.
Given a batch of training data and current NMT model parameterized by ${\theta}$, we repeatedly conduct forward propagation $K$ times\footnote{We empirically set $K$ to 5 following~\citet{self_paced_2020}.}.
On the $k$-th propagation, part of neurons $\hat{{\theta}}^{(k)}$ in network ${\theta}$ are randomly deactivated. 
Eventually, we obtain $K$ sets of model parameters $\{ \hat{{\theta}}^{(k)} \}_{k=1}^{K}$ and corresponding translation probabilities.
We use the expectation or variance of translation probabilities to estimate the model confidence \cite{uncertainty_bt_2019}.
Intuitively, the higher expectation or, the lower variance of translation probabilities reflects higher model confidence.
Formally at the $t$-th token position, we estimate the model confidence $conf(t)$ that calculated by the expectation of translation probabilities:
\begin{align}
conf(t) = \mathbb{E}\left[P(y_{t} | \mathbf{y}_{<t}, \mathbf{X}, \hat{\theta}^{(k)})\right]_{k=1}^{K}
\numberthis
\label{equ:mean}
\end{align}
We also use the variance of translation probabilities to estimate the model confidence $conf(t)$ as an alternative:
\begin{align}
conf(t) = 1 - \operatorname{Var}\left[P(y_{t} | \mathbf{y}_{<t}, \mathbf{X}, {\theta})\right]_{k=1}^{K}
\numberthis
\label{equ:var}
\end{align}
where $\operatorname{Var}[\cdot]$ denotes the variance of a distribution that calculated following the setting in \cite{uncertainty_bt_2019,uncertainty_cl_2020}.
We will further analyze the effect of different confidence estimations in Section \ref{sec:hyperparameter_experiments}.

\subsection{Confidence-Aware Scheduled Sampling}
The confidence score $conf(t)$ quantifies whether the current NMT model is confident
or hesitant on predicting the $t$-th target token.
We take $conf(t)$ as exact and real-time information to conduct a fine-grained schedule strategy in each training iteration. Specifically, a lower $conf(t)$ indicates that the current model $\theta$ still struggles with the teacher forcing mode for the $t$-th target token, namely underfitting for the conditional probability $P(y_{t} | \mathbf{y}_{<t}, \mathbf{X}, \theta)$. Thus we should keep feeding ground-truth tokens for learning to predict the $t$-th target token. Conversely, a higher $conf(t)$ indicates the current model $\theta$ has learned well the basic conditional probability under teacher forcing. Thus we should empower the model with the ability to cope with the exposure bias problem. Namely, we take inevitably erroneous model predictions as target inputs for learning to predict the $t$-th target.

Formally, in the second-pass decoding, the above fine-grained schedule strategy is conducted at all decoding steps simultaneously:
\begin{align}
y_{t-1} =
\begin{cases}
y_{t-1} \text{\ \ \ \ \ if \  $conf(t)  \leq t_{golden} $}  \\
\hat{y}_{t-1} 
  \text{\ \ \ \ \ else} 
\end{cases}
\numberthis
\label{equ:two_interval}
\end{align}
where $t_{golden}$ is a threshold to measure whether $conf(t)$ is high enough  ({\em e.g.,} 0.9) to sample the predicted token $\hat{y}_{t-1}$.

\subsection{Confidence-Aware Scheduled Sampling with Target Denoising}
Considering predicted tokens are obtained from the teacher forcing model, most predicted tokens ({\em e.g.,} about 70\% tokens  in WMT14 EN-DE) are the same as ground-truth tokens, which degenerate the scheduled sampling to the original teacher forcing. 
Although previous study~\cite{zhang_bridging_2019} have proposed to address this issue by using predictions from beam search, it conducts very slowly (about 4$\times$ slower than ours) due to the autoregressive property of beam search decoding.
To avoid the above degeneration problem while preserving computational efficiency, we try to add more noisy tokens instead of predicted tokens for high-confidence positions.
Inspired by~\citet{wechat_wmt_2020}, we replace ground-truth $y_{t-1}$ with a random token $y_{rand}$ of the current target sentence, which can simulate wordy and incorrect word order phenomena that occur at inference.
Considering $y_{rand}$ is more difficult\footnote{Given a pre-trained Transformer$_{base}$ model, we respectively replace ground-truth tokens with predicted tokens $\hat{y}$ or random tokens $y_{rand}$ with the same rate, and measure such difficulty by the increment of model perplexity. We observe that $y_{rand}$ yields about 15\% higher model perplexity than $\hat{y}$.} to learn than $\hat{y}_{t-1}$, we only adopt the noisy $y_{rand}$ for higher confidence positions.
Therefore, the fine-grained schedule strategy in Equation \ref{equ:two_interval} is extended to:
\begin{align}
y_{t-1} =
\begin{cases}
y_{t-1} \text{\ \ \ \ \ if \  $conf(t)  \leq t_{golden} $}  \\
\hat{y}_{t-1} 
  \text{\ \ \ \ \ if \ $t_{golden} < conf(t) \leq t_{rand} $} \\
 y_{rand} 
 \text{\ \ \ \ if \  $conf(t) > t_{rand} $}
\end{cases}
\numberthis
\label{equ:three_intervals}
\end{align}
where $t_{rand}$ is a threshold to measure whether $conf(t)$ is high enough  ({\em e.g.,} 0.95) to sample the random target token ${y}_{rand}$.
We provide detailed selections about  $t_{golden}$ and  $t_{rand}$ in Section \ref{sec:hyperparameter_experiments}.

\begin{table}[t!]
\begin{center}
\scalebox{0.90}{
\begin{tabular}{c c c}
\hline \textbf{Dataset} & \textbf{Size (M)} & \textbf{Valid / Test set} \\
\hline
WMT14 EN-DE &  4.5 & newstest 2013 / 2014 \\
WMT14 EN-FR &  36 & newstest 2013 / 2014 \\
WMT19 ZH-EN &  20 & newstest 2018 / 2019 \\
\hline
\end{tabular}}
\end{center}
\caption{Dataset statistics in our experiments.}
\label{table:dataset_statistics.}
\end{table}

\section{Experiments}
\label{sec:experiments}
We conduct experiments on three large-scale WMT 2014 English-German (EN-DE), WMT 2014 English-French (EN-FR), and WMT 2019 Chinese-English (ZH-EN) translation tasks.
We respectively build a shared source-target vocabulary for the EN-DE and EN-FR datasets, and unshared vocabularies for the ZH-EN dataset.
We apply byte-pair encoding \cite{bpe_2016} with 32k merge operations for all datasets. 
More datasets statistics are listed in Table \ref{table:dataset_statistics.}.

\subsection{Implementation Details}
\label{subsec:imp_details}
\paragraph{Training Setup.}
We train the Transformer$_{base}$ and Transformer$_{big}$ models \cite{transformer_2017}  with the open-source THUMT \cite{thumt_2017}.
All Transformer models are first trained by teacher forcing with 100k steps, and then trained with different training objects or scheduled sampling approaches for 300k steps. 
All experiments are conducted on 8 NVIDIA Tesla V100 GPUs, where each is allocated with a batch size of approximately 4096 tokens.
We use Adam optimizer \cite{Adam_2014} with 4000 warmup steps.
During training and the Monte Carlo Dropout process, we set dropout \cite{dropout_2014} rate to 0.1 for the Transformer$_{base}$ and 0.3 for the Transformer$_{big}$.

\paragraph{Evaluation.}
 We set the beam size to 4 and the length penalty to 0.6 during inference. We use \textit{multibleu.perl} to calculate case-sensitive BLEU scores for WMT14 EN-DE and EN-FR, and use \textit{mteval-v13a.pl} to calculate case-sensitive BLEU scores for WMT19 ZH-EN. 
We use the paired bootstrap resampling methods \cite{significance_test_2004} to compute the statistical significance of test results.

\subsection{Hyperparameter Experiments}
In this section, we elaborate hyperparameters settings involved in our approaches according to the performance on the validation set of WMT14 EN-DE, and share these settings for all WMT tasks.

\label{sec:hyperparameter_experiments}
\begin{table}[t!]
\begin{center}
\scalebox{0.88}{
\begin{tabular}{l c c c}
\hline \textbf{Methods} & \textbf{Training Cost} & \textbf{ BLEU } & \textbf{ $\Delta$ } \\ 
\hline
Transformer$_{base}$ & ref.   & 27.10 & ref. \\
\ \ \  $+$ PTP & 1.3$\times$      & 28.15 & +1.05 \\
\ \ \  $+$ Expectation & 2.7$\times$  & 28.15 & +1.05 \\
\ \ \  $+$ Variance & 2.7$\times$  & 28.20 & +1.10 \\
\hline
\end{tabular}}
\end{center}
\caption{BLUE scores (\%) on the validation set of WMT14 EN-DE with different confidence estimations. `Training Cost' is calculated by the total training time until models convergence on 8 NVIDIA V100 GPUs.
`PTP' refers to PTP-based confidence estimation in Equation (\ref{equ:ptp}).
`Expectation'  and `Variance' refers to Monte Carlo dropout sampling-based confidence estimation in Equation (\ref{equ:mean}) and (\ref{equ:var}), respectively.
`ref.' is short for the reference baseline.
}
\label{table:different_confidence_estimation}
\end{table}

\paragraph{Different Confidence Estimations.}
In this section, we analyze effects of different estimations for model confidence described in Section \ref{sec:model_confidence_estimation}. 
As shown in Table \ref{table:different_confidence_estimation}, we observe that  Monte Carlo dropout sampling based approaches ({\em i.e.,} expectation and variance of translation probabilities)  achieve  comparable or  marginally better translation quality than PTP. However,  since  Monte Carlo dropout sampling based approaches  need additional passes for forward propagation, which yields about 2.7$\times$ computation costs than the Transformer$_{base}$. On the contrary, PTP only causes marginal additional computation costs (1.3$\times$) than the Transformer$_{base}$, as PTP can be directly obtained in the first pass decoding.
Considering the trade-off between training efficiency and final performance, we use PTP to estimate model confidence by default in the following experiments.

\begin{figure}[t!]
\begin{center}
     \scalebox{0.45}{
      \includegraphics[width=1\textwidth]{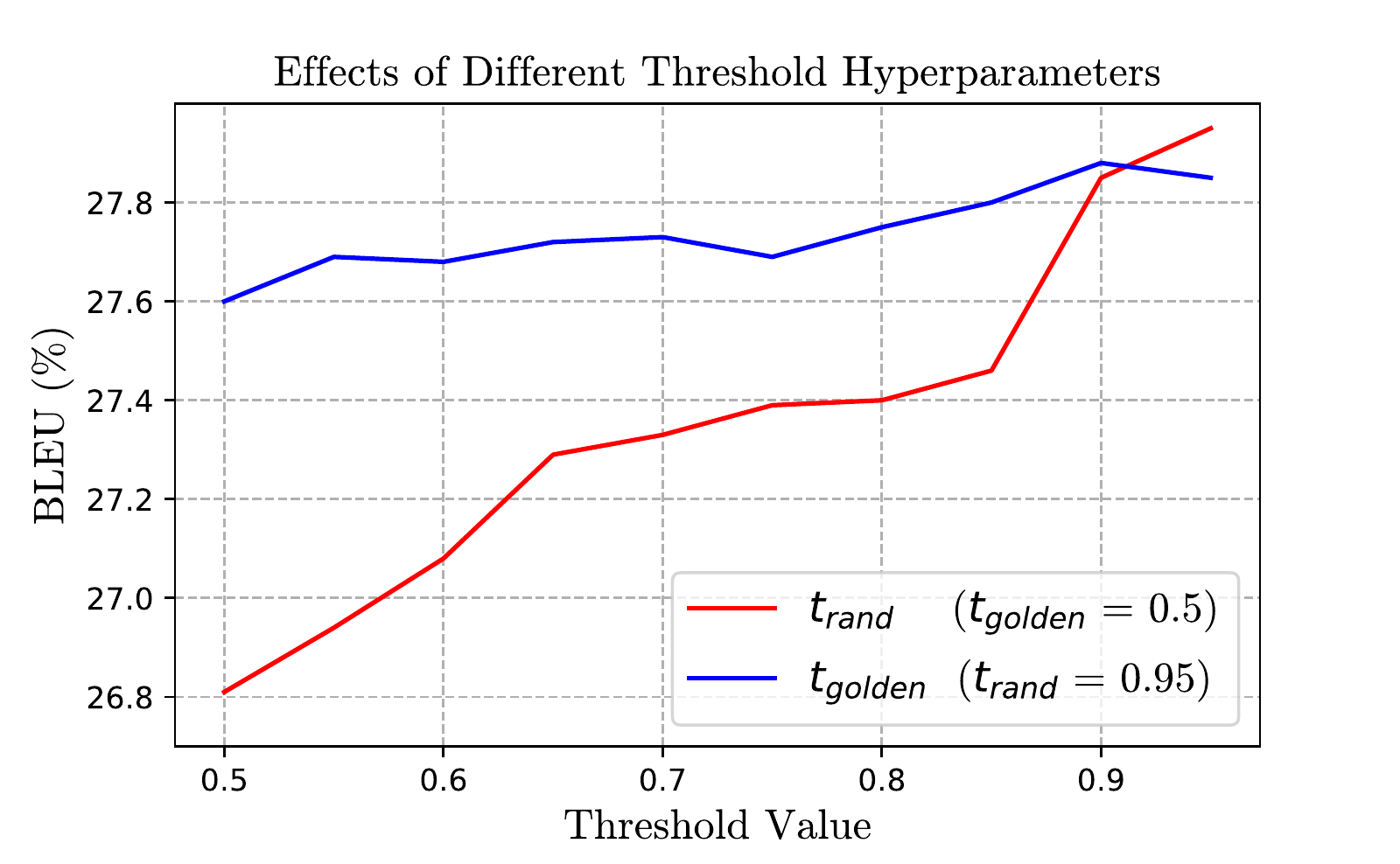}
      } 
      \caption{BLUE scores (\%) on the validation set of WMT14 EN-DE with different $t_{golden}$ and $t_{rand}$.
      } 
      \label{fig:conf_hyperparams}  
 \end{center} 
\end{figure}

\paragraph{Thresholds Settings.}
There are two important hyperparameters in our approaches, namely the two threshold $t_{golden}$ and $t_{rand}$ that determine token selections in Equation (\ref{equ:three_intervals}).
In our preliminary experiments, we observe our approach is relatively not sensitive to $t_{golden}$, thus we firstly fix  $t_{golden}$ to a modest value, {i.e.,} 0.5 and analyze effects when $t_{rand}$ ranging from 0.5 to 0.95. 
As the red line is shown in Figure \ref{fig:conf_hyperparams}, we observe that a rapid improvement in performance with the growth of $t_{rand}$. Therefore, we decide to set $t_{rand}$ to 0.95 and then analyze effects when $t_{golden}$ ranging from 0.5 to 0.95.  
As the blue line is shown in Figure \ref{fig:conf_hyperparams},  the model performance gently rises with the growth of $t_{golden}$ and finally achieves its peak when  $t_{golden} = 0.9$.  Thus we finally set $t_{golden}$ to 0.9.

\begin{table*}[t!]
\begin{center}
\scalebox{0.89}{
\begin{tabular}{l c c c c} 
\hline
\multirow{2}*{\textbf{Model}}  &  \multicolumn{3}{c}{\textbf{BLEU}}  \\ \cline{2-4}
~ & \textbf{EN-DE} & \textbf{ZH-EN} & \textbf{EN-FR}  \\
\hline
Transformer$_{base}$ \cite{transformer_2017} & 27.30 & -- & 38.10   \\
Transformer$_{base}$ \cite{transformer_2017} $\dagger$ & 27.90 & 24.97 & 40.30   \\
\ \ \  $+$ Mixer \cite{mixer_2015} $\dagger$  & 28.54 & 25.28 & 40.57 \\
\ \ \  $+$ Minimal Risk Training  \cite{mrt_2016} $\dagger$   & 28.55 & 25.23 & 40.82 \\
\ \ \  $+$ TeaForN \cite{teaforn_2020}  & 27.90 & -- & 40.84   \\
\ \ \  $+$ TeaForN \cite{teaforn_2020} $\dagger$ & 28.60 & 25.45 & 40.94 \\
\ \ \  $+$ Self-paced learning \cite{self_paced_2020} $\dagger$  & 28.85 & 25.56 & 41.12 \\
\ \ \  $+$ Vanilla scheduled sampling \cite{ss_2015} $\dagger$  & 28.40 & 25.43 & 40.87 \\
\ \ \  $+$ Target denoising \cite{wechat_wmt_2020} $\dagger$ & 28.55 & 25.58 & 40.57 \\
\ \ \  $+$ Sampling with sentence oracles  \cite{zhang_bridging_2019} & 28.65 & --  & -- \\
\ \ \  $+$ Sampling with sentence oracles  \cite{zhang_bridging_2019} $\dagger$ & 28.65 & 25.50 & 40.85 \\
\ \ \  $+$ Confidence-aware scheduled sampling (ours) $\dagger$ & ~~28.80$*$ & ~~~~25.95$**$ & ~~~~41.19$**$ \\
\ \ \  $+$ Confidence-aware scheduled sampling with target denoising (ours) $\dagger$ & ~~~~\textbf{28.91}$**$ & ~~~~\textbf{26.00}$**$ & ~~~~\textbf{41.28}$**$ \\
\hline
Transformer$_{big}$ \cite{transformer_2017}  & 28.40 & -- & 41.80 \\
Transformer$_{big}$ \cite{transformer_2017} $\dagger$ & 28.90 & 25.22 & 41.89 \\
\ \ \  $+$ Mixer \cite{mixer_2015} $\dagger$ & 29.27 & 25.58 & 42.37 \\
\ \ \  $+$ Minimal Risk Training  \cite{mrt_2016} $\dagger$  & 29.35 & 25.65 & 42.46 \\
\ \ \  $+$ TeaForN \cite{teaforn_2020}  & 29.30 & -- & 42.73 \\
\ \ \  $+$ TeaForN \cite{teaforn_2020} $\dagger$  & 29.32 & 25.48 & 42.62 \\
\ \ \  $+$ Error correction \cite{error_correction_2020} & 29.20 & -- & -- \\
\ \ \  $+$ Self-paced learning \cite{self_paced_2020} $\dagger$ & 29.68 & 25.56 & 42.32 \\
\ \ \  $+$ Vanilla scheduled sampling \cite{ss_2015} $\dagger$ & 29.62 & 25.60 & 42.55 \\
\ \ \  $+$ Target denoising \cite{wechat_wmt_2020} $\dagger$  & 29.18 & 25.56 & 42.32 \\
\ \ \ $+$ Scheduled sampling with sentence oracles  \cite{zhang_bridging_2019} $\dagger$ & 29.57 & 25.78 & 42.65 \\
\ \ \  $+$ Confidence-aware scheduled sampling (ours) $\dagger$ & ~~~~29.95$**$ & ~~~~26.00$**$ & ~~~~42.90$**$ \\
\ \ \  $+$ Confidence-aware scheduled sampling with target denoising (ours) $\dagger$ & ~~~~\textbf{30.09}$**$ & ~~~~\textbf{26.27}$**$ & ~~~~\textbf{42.97}$**$ \\
\hline
\end{tabular}}
\end{center}
\caption{Translation performance on each WMT dataset. `$\dagger$' is our implementations under unified settings. The original TeaForN \cite{teaforn_2020} reports SacreBLEU scores. For fair comparison, we re-implement it and report BLEU scores.
`$*$/$**$': significantly \cite{significance_test_2004} better than `Vanilla Scheduled Sampling'  with $p < 0.05$ and $p < 0.01$.
}
\label{table:nmt_results}
\end{table*}

\subsection{Systems}
\paragraph{Mixer.}
A sequence-level training algorithm for text generations by combining both REINFORCE and cross-entropy \cite{mixer_2015}.

\paragraph{Minimal Risk Training.}
Minimal Risk Training (MRT) \cite{mrt_2016} introduces evaluation
metrics ({\em e.g.,} BLEU) as loss functions and aims to minimize expected loss on the training data.

\paragraph{TeaForN.}
Teacher forcing with n-grams \cite{teaforn_2020} enable the standard teacher forcing with a broader view by n-grams optimization.

\paragraph{Self-paced learning.}
\citet{self_paced_2020} assign confidence
scores for each input to weight its loss.

\paragraph{Vanilla schedule sampling.}
Scheduled sampling on training steps with the inverse sigmoid decay~\cite{ss_2015,zhang_bridging_2019}.

\paragraph{Sampling with sentence oracles.}
\citeauthor{zhang_bridging_2019} (\citeyear{zhang_bridging_2019}) refine the sampling space of scheduled sampling with sentence oracles, {\em i.e.,} predictions from beam search. Note that its sampling strategy is still based on training steps with the sigmoid decay.

\paragraph{Target denoising.}
\citeauthor{wechat_wmt_2020} (\citeyear{wechat_wmt_2020}) add noisy perturbations into decoder inputs when training, which yields a more robust translation model against prediction errors by target denoising.

\paragraph{Confidence-aware scheduled sampling.}
Our fine-grained schedule strategy described in Equation (\ref{equ:two_interval}) with $t_{golden}=0.9$. 

\paragraph{Confidence-aware scheduled sampling with target denoising.}
Our fine-grained schedule strategy described in Equation (\ref{equ:three_intervals}) with $t_{golden}=0.9$ and $t_{rand}=0.95$ .

\begin{table}[t!]
\begin{center}
\scalebox{0.90}{
\begin{tabular}{l c c}
\hline \textbf{Schedule Strategy} & \textbf{ BLEU } & \textbf{ $\Delta$ } \\ 
\hline
Transformer$_{base}$  & 27.10  &  ref. \\
\ \ \  $+$ Linear decay & ~~27.56$*$ & +0.46 \\
\ \ \  $+$ Exponential decay & ~~27.60$*$ & +0.50 \\
\ \ \  $+$ Inverse sigmoid decay  & ~~27.65$*$ & +0.55 \\
\ \ \  $+$ Confidence (ours) & ~~~~\textbf{28.15}$**$ & +\textbf{1.05} \\
\hline
\end{tabular}}
\end{center}
\caption{BLUE scores (\%) on the validation set of WMT14 EN-DE with different schedule strategies.
`Confidence' refers to the confidence-aware strategy in Equation (\ref{equ:two_interval}).
`ref.' is short for the reference baseline.
`$*$ / $**$': significantly \cite{significance_test_2004} better than the Transformer$_{base}$ with $p < 0.05$ and $p < 0.01$.
}
\label{table:effectiveness_of_confidence}
\end{table}

\subsection{Main Results}
We list translation qualities in Table \ref{table:nmt_results}.
For the Transformer$_{base}$ baseline, our `Confidence-aware scheduled sampling' shows consistent improvements by 0.90, 0.98, 0.89 BLEU points on EN-DE, ZH-EN, and EN-FR, respectively. Moreover, after applying the more fine-grained strategy with target denoising, our `Confidence-aware scheduled sampling with target denoising' achieves further improvements which are 1.01, 1.03, 0.98 BLEU points on EN-DE, ZH-EN, and EN-FR, respectively.
When comparing with the stronger vanilla scheduled sampling method, `Confidence-aware scheduled sampling with target denoising' still yields improvements by 0.51, 0.57, and 0.41 BLEU points on the above three tasks, respectively.
For the more powerful Transformers$_{big}$, we also observe similar experimental conclusions as above. Specifically, `Confidence-aware scheduled sampling with target denoising' outperforms vanilla scheduled sampling by 0.47, 0.67, and 0.42 BLEU points, respectively.
In summary,  experiments on strong baselines and various tasks verify the effectiveness and  superiority of our approaches.

\section{Analysis and Discussion}
We analyze our proposals on WMT 2014 EN-DE with the Transformer$_{base}$ model.

\subsection{Effects of Confidence-Aware Strategies}
\label{sec:effectiveness_of_confidence}
In this section, we rigorously validate the effectiveness of confidence-aware strategies by univariate experiments with the only difference at schedule strategy.
As shown in Table \ref{table:effectiveness_of_confidence}, existing heuristic functions, {\em i.e.,} linear, exponential, and inverse sigmoid decay, moderately bring improvements over the Transformer$_{base}$ baseline by 0.46, 0.50, and 0.55 BLEU points, respectively. While our confidence-aware strategy that described in Equation (\ref{equ:two_interval}) can significantly outperform the baseline by 1.05 BLEU points.
We attribute the effectiveness of the confidence-aware strategy to its exact and suitable token assignments according to the real-time model competence rather than predefined patterns.

\begin{table}[t!]
\begin{center}
\scalebox{0.89}{
\begin{tabular}{l c c}
\hline \textbf{Model} & \textbf{ BLEU } & \textbf{ $\Delta$ } \\ 
\hline
Our approach  & 28.15  & ref. \\
\ \ \  $-$ Confidence & 27.75 & -0.40 \\
\ \ \  $-$ Denoising & 28.00 & -0.15 \\
\ \ \  $-$ Confidence \& Denoising & 27.64 & -0.51 \\
\hline
\end{tabular}}
\end{center}
\caption{BLUE scores (\%) on the validation set of WMT14 EN-DE for ablation experiments.
`Our approach' is `confidence-aware scheduled sampling with target denoising' in Equation (\ref{equ:three_intervals}).
`Confidence' refers to the confidence-aware strategy in Equation (\ref{equ:three_intervals}).
`Denoising' refers to the target random noise $y_{rand}$ in Equation (\ref{equ:three_intervals}).
`ref.' is short for the reference baseline.
}
\label{table:ablation}
\end{table}

\subsection{Ablation Experiments} 
We conduct ablation experiments to investigate the impacts of various components in our `Confidence-aware scheduled sampling with target denoising' (described in Equation (\ref{equ:three_intervals})) and list results in Table \ref{table:ablation}.
Separately removing the confidence-aware strategy degenerates our approach into the vanilla target denoising with a uniform strategy \cite{wechat_wmt_2020}, which causes a noticeable drop (0.4 BLEU), indicating the confidence-aware strategy plays a leading role for performance. 
On the other hand, we only observe a drop (0.15 BLEU) when removing `Target denoising', revealing the additional noise plays a secondary role for performance.
Finally, ablating both the confidence-aware strategy and `Target denoising' degenerates our approach into the vanilla scheduled sampling. It yields a further decrease (0.51 BLEU), suggesting the confidence-aware strategy and `Target denoising' are complementary with each other.

\begin{figure}[t!]
\begin{center}
     \scalebox{0.47}{
      \includegraphics[width=1\textwidth]{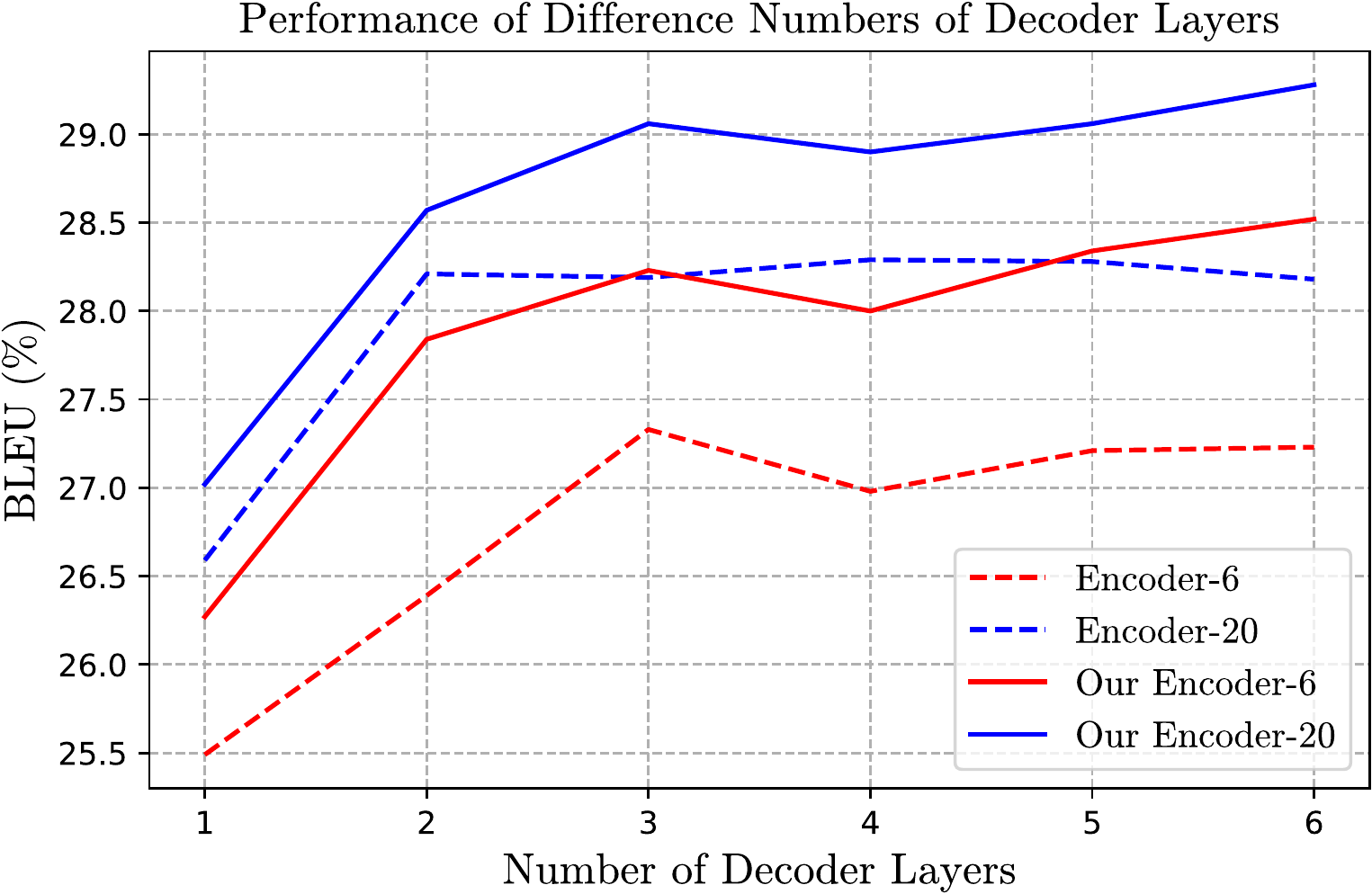}
      } 
      \caption{
      BLUE scores (\%) on the validation set of WMT14 EN-DE with different numbers of decoder layers. Solid lines refer to our confidence-aware schedule strategy. Dashed lines refer to the Transformer$_{base}$.
      } 
      \label{fig:difference_model_capacities}  
 \end{center} 
\end{figure}

\subsection{Different Numbers of Decoder Layers}
As known in existing studies \cite{deep_transformer_2018,deep_transformer_2019}, there exists a performance bottleneck at the decoder side of NMT models. Namely, the increase in the number of decoder layers can not bring corresponding improvements for performance. 
\citet{easy_but_sensitive_2019} attribute this bottleneck to the fact that decoders learn an easier task than encoders.

In this paper, our fine-grained schedule strategy in Equation (\ref{equ:three_intervals}) assigns a more difficult task to the decoder.
We can not help wondering whether our strategy is able to alleviate the above performance bottleneck.
Firstly, we keep the number of encoders fixed to 6 ({\em i.e.,} Encoder-6), then apply our confidence-aware schedule strategy on the Encoder-6 Transformer$_{base}$ with the number of decoder layers ranging from 1 to 6.
As shown in Figure \ref{fig:difference_model_capacities}, our approach (solid red line) consistently outperforms the Encoder-6 Transformer$_{base}$ (dashed red line). 
More importantly, the improvement of Encoder-6 Transformer$_{base}$ stops ({\em i.e.,} performance bottleneck) once the number of decoder exceeds 4.
Despite this, we observe continuous improvement with the growth of decoder layers in our approach.
Moreover, we repeat the above experiments with more powerful deep encoders (Encoder-20). 
We observe that the performance bottleneck for Encoder-20 Transformer$_{base}$ becomes more evident (dashed blue line).
Despite this, our approaches (solid blue line) still keep improving performance with the growth of decoder layers on the stronger Encoder-20 Transformer$_{base}$.

In summary, our confidence-aware schedule strategy brings a meaningful increase in the difficulty of decoders, and the bottleneck at the decoder side is alleviated to a certain extend.

\begin{figure}[t!]
\begin{center}
     \scalebox{0.48}{
      \includegraphics[width=1\textwidth]{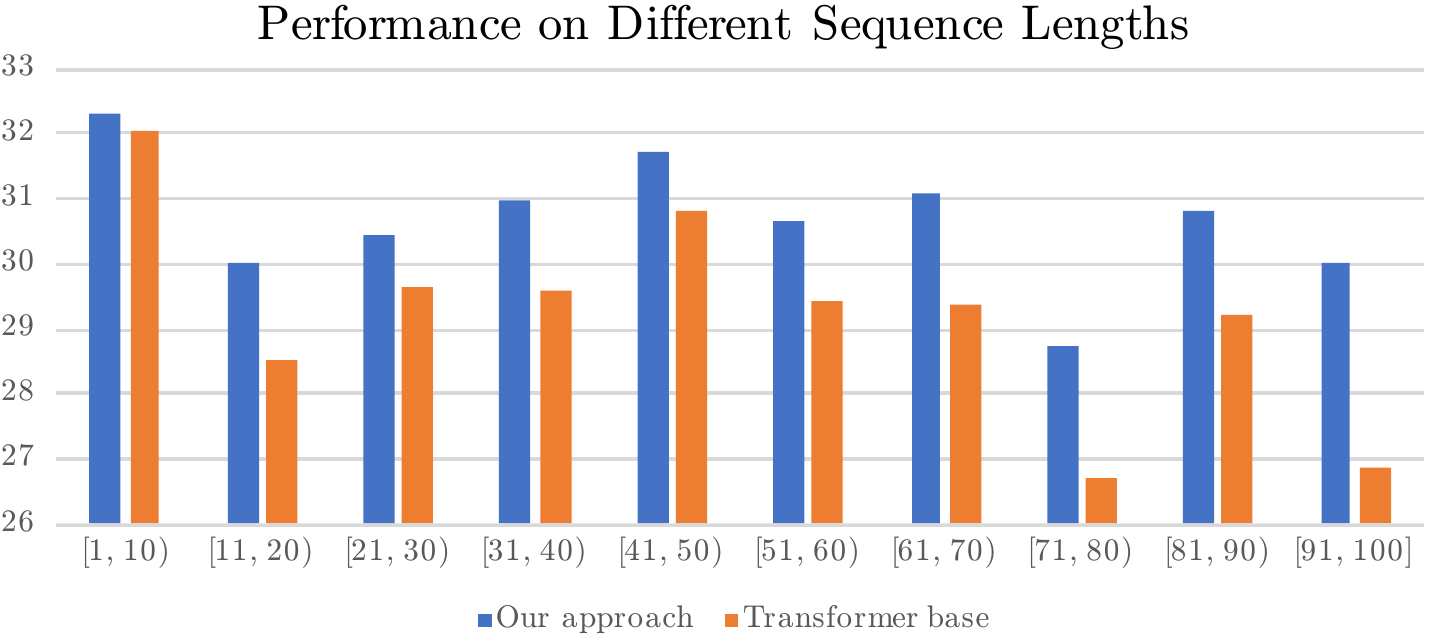}
      } 
      \caption{BLUE scores (\%) on the randomly sampled  WMT14 EN-DE training data with different lengths. 
      } 
      \label{fig:performance_on_different_sequence_lengths}  
 \end{center} 
\end{figure}

\subsection{Effects on Different Sequence Lengths}
Due to error accumulations, the exposure bias problem becomes more problematic with the growth of sequence lengths \cite{sdnmt_2019,bi_direction_2020}. 
Thus it is intuitive to verify the effectiveness of our approach over different sequence lengths.
Considering the validation set of WMT14 EN-DE (3k) is too small to cover scenarios with various sentence lengths, we randomly select 10k training data with lengths from 10 to 100. 
As shown in Figure \ref{fig:performance_on_different_sequence_lengths}, our approach consistently outperform the Transformer$_{base}$ model at different sequence lengths. Moreover, the improvements of our approach over the Transformer$_{base}$ is gradually increasing with sentence lengths. Specifically, we observe more than 1.0 BLEU improvements when sentence lengths in $[80, 100]$.

\subsection{Model  Convergence}
\label{sec:model_convergence}
\begin{figure}[t!]
\begin{center}
     \scalebox{0.47}{
      \includegraphics[width=1\textwidth]{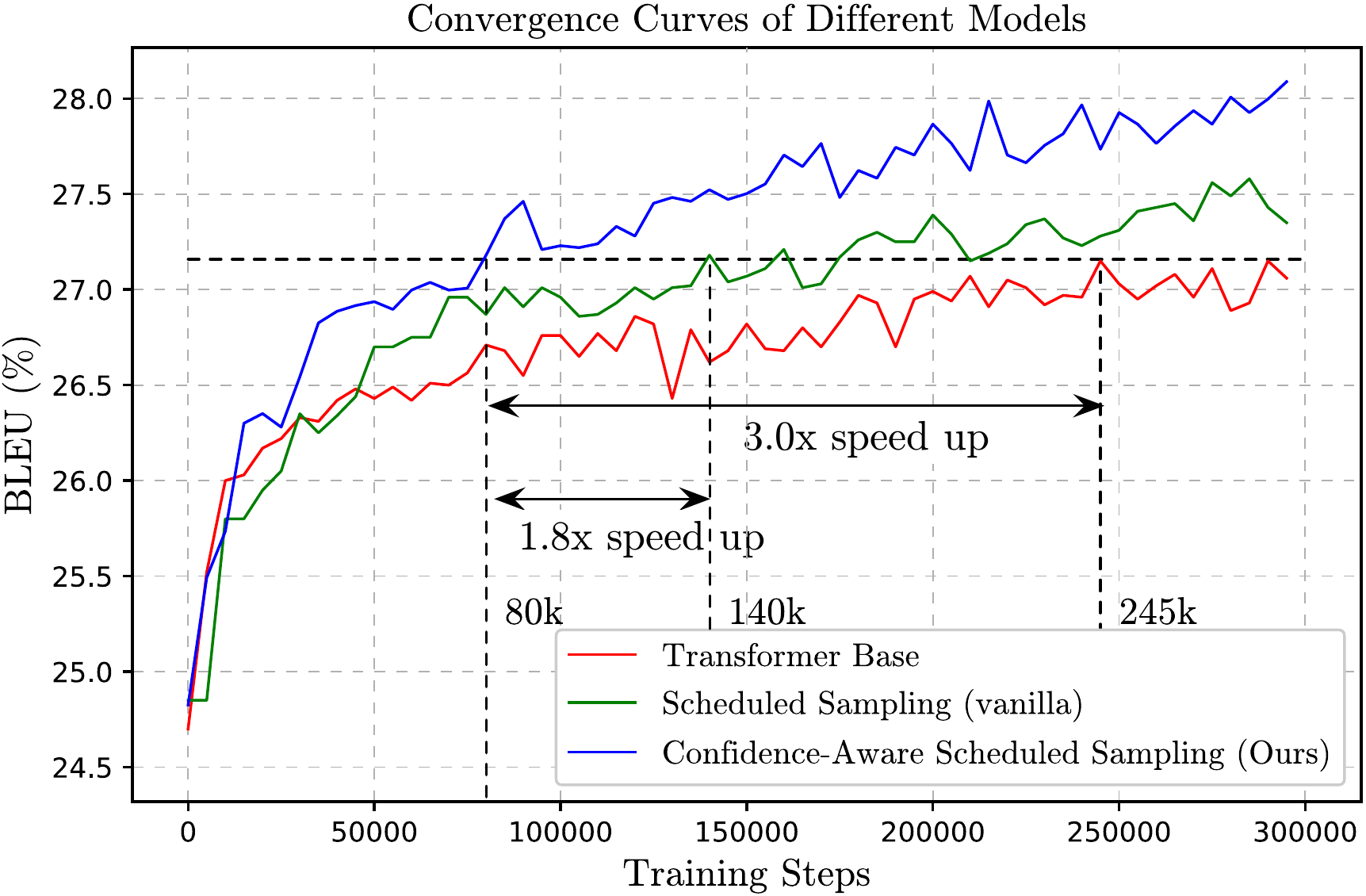}
      } 
      \caption{Convergence curves for different models.
      BLUE scores (\%) are calculated on the validation set of WMT14 EN-DE.
      Our approach can achieve the same performance as the Transformer$_{base}$ with about 3.0$\times$ speed up.
      } 
      \label{fig:converage_curve}  
 \end{center} 
\end{figure}

As aforementioned, our confidence-aware scheduled sampling learns to deal with the exposure bias problem in an efficient manner, thus speeding up the model convergence.
As shown in Figure \ref{fig:converage_curve}, it costs the Transformer$_{base}$ 245k steps to converge to a local optimum (about 27.1 BLEU). To achieve the same performance, it only costs our confidence-aware scheduled sampling 80k step, namely about 3.0$\times$ speed up over the Transformer$_{base}$ and 1.8$\times$ speed up over the vanilla scheduled sampling.
Since vanilla scheduled sampling randomly exposes more difficult predicted tokens for each token position, regardless of the actual model competence, its convergence speed is restricted to a certain extent. 
On the contrary, our approach samples predicted tokens only if the current model is capable of dealing with these more difficult inputs, mimicking the learning process of humans.
Therefore, our approach is trained more efficiently.


\section{Related Work}

\paragraph{Confidence-aware Learning for NMT.}
As to confidence estimations for NMT, \citet{nmt_confidence_2015} frame translation as a compression game and measure the amount of information added by translators. \citet{uncertainty_bt_2019} propose to quantify the confidence of NMT model predictions based on model uncertainty, which is widely extend to select training samples~\cite{data_selection_2020,dynamic_bt_2020}, to design confidence-aware curriculum learning \cite{uncertainty_cl_2020,self_paced_2020},  and to  augment synthetic corpora \cite{uncertainty_augmentation_2020}. Model confidence is also served as a useful metric for analyze NMT model from the perspective of fitting and search \cite{analyze_uncertainty_nmt_2018}, visualization \cite{visualizing_confidence_2017} and  calibration \cite{calibration_2019,calibration_acl_2020}.  Different from existing studies, we are the first to propose confidence-aware scheduled sampling for alleviating the exposure bias problem in NMT.

\section{Conclusion}
 In this paper, we propose confidence-aware scheduled sampling for NMT, which exactly samples corresponding tokens according to the real-time model competence rather than human intuitions. We further explore to sample more noisy tokens for high-confidence token positions, preventing scheduled sampling from degenerating into the original teacher forcing mode. Experiments on three large-scale WMT translation tasks suggest that our approach improves vanilla scheduled sampling both translation quality and convergence speed. We elaborately analyze the effectiveness and efficiency of our approach from multiple aspects.
 As a result, we further observe our approaches: 1) can alleviate the performance bottleneck of decoders for NMT to a certain extend; 2) improve the translation quality of long sequences. 
 
 \section*{Acknowledgments}
The research work descried in this paper has been supported by the National Key R\&D Program of China (2020AAA0108001) and the National Nature Science Foundation of China (No. 61976015, 61976016, 61876198 and  61370130). The authors would like to thank the anonymous reviewers for their valuable comments and suggestions to improve this paper.

\bibliography{acl2021}
\bibliographystyle{acl_natbib}
\end{document}


\maketitle

\paragraph{A: Decay Strategies on Training Steps}

\begin{figure}[t!]
\begin{center}
     \scalebox{0.5}{
      \includegraphics[width=1\textwidth]{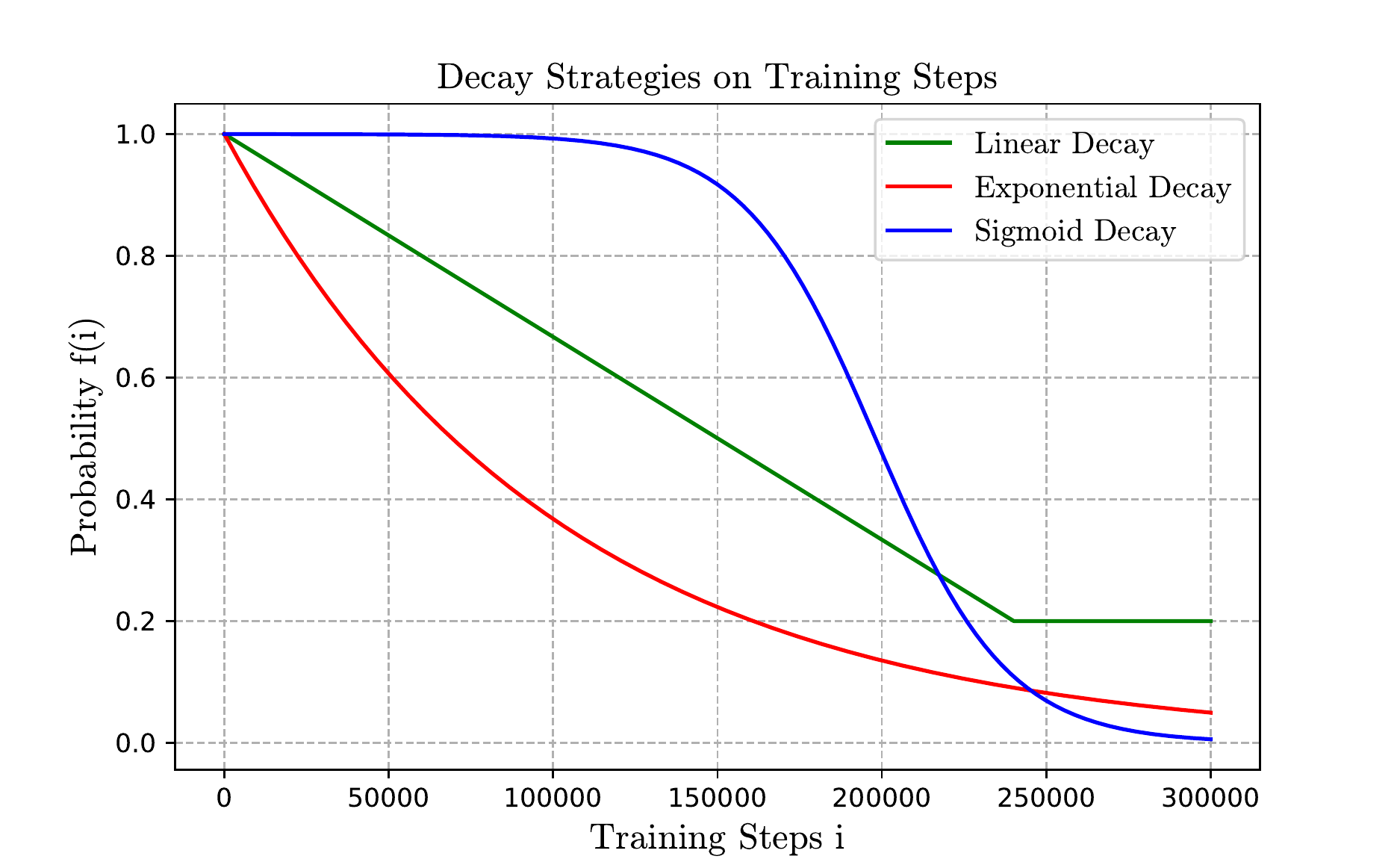}
      } 
      \caption{
      Examples of different decay strategies $f(i)$.
      } 
      \label{fig:training_step_decay_strategy}  
 \end{center} 
\end{figure}

\label{sec:decay_strategies_on_training_steps}
Existing schedule strategies are based on training steps \cite{ss_2015,zhang_bridging_2019}. As the number of the training step $i$ increases, the model should be exposed to its own predictions more frequently.  At the $i$-th training step, the probability of sampling golden tokens $f(i)$ is calculated as follow:
\begin{itemize}
\item Linear Decay: $f(i) = \max(\epsilon , ki + b)$, where $\epsilon$ is the minimum value, and $k < 0$ and $b$ are respectively the slope and offset of the decay.
\item Exponential Decay: $f(i) = k^i$, where $k < 1$ is the radix to adjust the sharpness of the decay.
\item Inverse Sigmoid Decay: $f(i) = \frac{k}{k + e^{\frac{i}{k}}}$, where $e$ is the mathematical constant, and $k \geq 1$ is a hyperparameter to adjust the sharpness of the decay.
\end{itemize}
To reproduce the vanilla scheduled sampling, we slightly tune hyperparameters for different decay strategies according to the performance on the validation set of WMT14 EN-DE.
For the linear decay, we set $\epsilon$, $k$, and  $b$ to 0.2, -5e-5, and 1, respectively.
For the exponential decay,  we set $k$ to 0.99999.
For the inverse sigmoid decay, we set $k$ to 20,000.
We draw visible examples for different decay strategies with the above hyperparamethers in Figure \ref{fig:training_step_decay_strategy}.

\bibliography{acl2021}
\bibliographystyle{acl_natbib}